\documentclass[10pt,twocolumn,letterpaper]{article}

\usepackage{wacv}
\usepackage{times}
\usepackage{epsfig}
\usepackage{graphicx}
\usepackage{amsmath}
\usepackage{amssymb}
\usepackage{placeins}
\usepackage{multirow}
\usepackage[skip=5pt]{caption}
\usepackage{booktabs}
\usepackage{rotating}
\usepackage{booktabs}
\usepackage{cancel}
\usepackage{setspace}
\usepackage{eso-pic}
\usepackage[accsupp]{axessibility}  % Improves PDF readability for those with disabilities.

\wacvfinalcopy
\renewcommand{\thefootnote}{\alph{footnote}}
 
\def\wacvPaperID{736} % Enter the WACV Paper ID here

\ifwacvfinal
\fi

\ifwacvfinal
\usepackage[breaklinks=true,bookmarks=false]{hyperref}
\else
\usepackage[pagebackref=true,breaklinks=true,colorlinks,bookmarks=false]{hyperref}
\fi

\newcommand{\fig}[1]{Figure~\ref{fig:#1}}

\newcommand{\tab}[1]{Table~\ref{tab:#1}}

\begin{document}

%%%%%%%%% TITLE
\title{Resolution-robust Large Mask Inpainting with Fourier Convolutions}

\author{Roman Suvorov$^1$~~~Elizaveta Logacheva$^1$~~~Anton Mashikhin$^1$~~~Anastasia Remizova$^{3*}$~~~Arsenii Ashukha$^1$\\
Aleksei Silvestrov$^1$~~~Naejin Kong$^2$~~~Harshith Goka$^2$~~~Kiwoong Park$^2$~~~Victor Lempitsky$^{1,4}$\\
\vspace{-5pt}\\
\normalsize $^1$\textsl{Samsung AI Center Moscow}, $^2$\textsl{Samsung Research},\\
\normalsize $^3$\textsl{School of Computer and Communication Sciences, EPFL}, \\
\normalsize $^4$\textsl{Skolkovo Institute of Science and Technology, Moscow, Russia}
}

\twocolumn[{
    \vspace{-25pt}
    \renewcommand\twocolumn[1][]{#1}
    \maketitle
    \centering

    \vspace{-20pt}
    \includegraphics[width=1\textwidth]{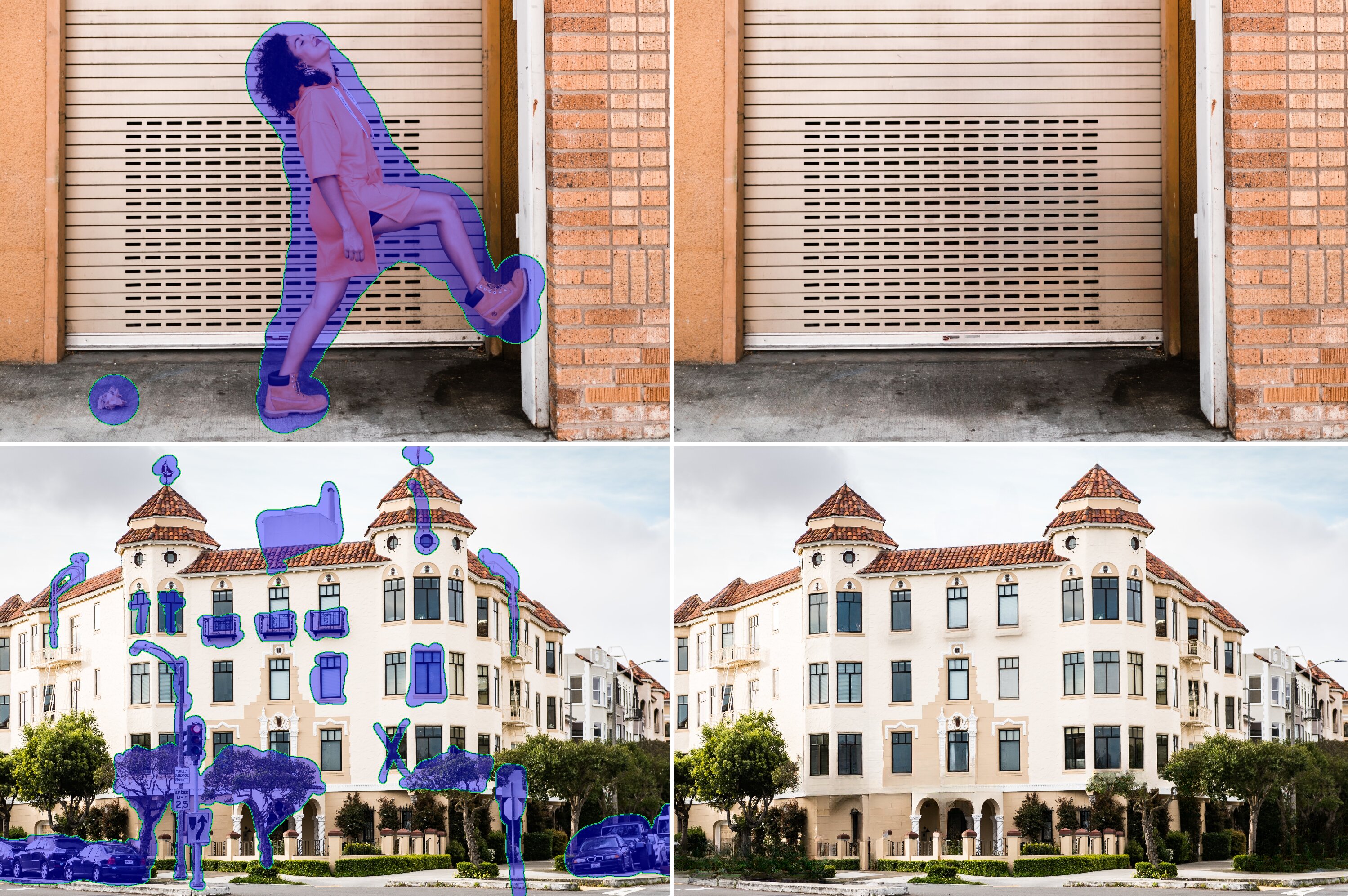}

    \captionof{figure}{
    The proposed method can successfully inpaint large regions and works well with a wide range of images, including those with complex repetitive structures. 
    The method generalizes to high-resolution images, while trained only in low $256\times256$ resolution.}
    \label{fig:teaser}
}]

\maketitle

\newcommand\blfootnote[1]{%
  \begingroup
  \renewcommand\thefootnote{}\footnote{#1}%
  \addtocounter{footnote}{-1}%
  \endgroup
}

\thispagestyle{empty}
\begin{abstract}
\vspace{-10pt}
\footnotetext{Correspondence to Roman Suvorov {\tt windj007@gmail.com}}
\footnotetext{$^*$ The work is done while at Samsung AI Center Moscow}
\setcounter{footnote}{0}
Modern image inpainting systems, despite the significant progress, often struggle with large missing areas, complex geometric structures, and high-resolution images. We find that one of the main reasons for that is the lack of an effective receptive field in both the inpainting network and the loss function. To alleviate this issue, we propose a new method called large mask inpainting (LaMa). LaMa is based on $i)$~a new inpainting network architecture that uses fast Fourier convolutions~(FFCs), which have the image-wide receptive field; $ii)$~a high receptive field perceptual loss; $iii)$~large training masks, which unlocks the potential of the first two components. Our inpainting network improves the state-of-the-art across a range of datasets and achieves excellent performance even in challenging scenarios, e.g. completion of periodic structures. Our model generalizes surprisingly well to resolutions that are higher than those seen at train time, and achieves this at lower parameter\&time costs than the competitive baselines. The code is available at {\footnotesize \url{https://github.com/saic-mdal/lama}}.
\end{abstract}

\definecolor{trolleygrey}{rgb}{0.5, 0.5, 0.5}
\definecolor{darkpastelgreen}{rgb}{0.01, 0.75, 0.24}
\definecolor{darkpink}{rgb}{0.91, 0.33, 0.5}
\definecolor{alizarin}{rgb}{0.82, 0.1, 0.26}
\definecolor{americanrose}{rgb}{1.0, 0.01, 0.24}
\vspace{-10pt}
\section{Introduction}
%\vspace{3pt}
The solution to the image inpainting problem---realistic filling of missing parts---requires both to ``understand`` large-scale structure of natural images and to perform image synthesis. 
The subject has been studied in pre-deep learning era~\cite{Bertalmio03texturestructure, Criminisi03exemplar,Hays07millions}, and the progress accelerated in recent years through the use of deep and wide neural networks~\cite{liu2018image, ma2019region, li2020recurrent} and adversarial learning~\cite{pathak2016context, iizuka2017globally, yu2018generative, song2018spg, yu2019free, nazeri2019edgeconnect, yi2020contextual, zeng2020high}. 
% In this work, we reassess several aspects of deep large-mask image inpainting, providing and validating several improvements, and pushing the state-of-the-art for this problem.

The usual practice is to train inpainting systems on a large automatically generated dataset, created by randomly masking real images. 
It's common to use complicated two-stage models with intermediate predictions, such as smoothed images~\cite{liu2020rethinking,yi2020contextual,zeng2020high}, edges~\cite{nazeri2019edgeconnect,xiong2019foreground}, and segmentation maps~\cite{song2018spg}. In this work, we achieve state-of-the-art results with a simple single-stage network. 

A large effective receptive field \cite{Luo16effective} is essential for understanding the global structure of an image and hence solving the inpainting problem. 
Moreover, in the case of a large mask, an even large yet limited receptive field may not be enough to access information necessary for generating a quality inpainting.
% Moreover, in case of a large mask, an even large yet limited receptive field may not allow a network to access necessary information---the network will observe only $0$-pixels inside the mask. 
We notice that popular convolutional architectures might lack a sufficiently large effective receptive field. We carefully intervene into each component of the system to alleviate the problem and to unlock the potential of the single-stage solution. Specifically:

% We noticed that commonly used convolutional architectures lack a sufficiently large receptive field---that is essential for the understanding of the global structure, and hence solving inpainting. To unlock the potential of the single-stage solution we carefully intervene in each component of the system: architecture, loss functions, and training data generation procedure. More specifically our contributions are:
\textbf{\textit{i)}} We propose an inpainting network based on recently developed \textit{fast Fourier convolutions}~(\textit{FFCs})~\cite{chi2020ffc}. FFCs allow for a receptive field that covers an entire image even in the early layers of the network.
%We show that the network benefits from this property of FFC, achieving both higher quality and parameter efficiency.
We show that this property of FFCs improves both perceptual quality and parameter efficiency of the network.
Interestingly, the inductive bias of FFC allows the network to generalize to high resolutions that are never seen during training~(\fig{res-grow}, \fig{high-res-degradation-quantitative}). This finding brings significant practical benefits, as less training data and computations are needed.

\textbf{\textit{ii)}} We propose the use of the perceptual loss~\cite{johnson2016perceptual} based on a semantic segmentation network with a high receptive field. This leans upon the observation that the insufficient receptive field impairs not only the inpainting network, but also the perceptual loss.  Our loss promotes the consistency of global structures and shapes.

\textbf{\textit{iii)}} We introduce an aggressive strategy for training mask generation, to unlock the potential of a high receptive field of the first two components. The procedure produces wide and large masks, which force the network to fully exploit the high receptive field of the model and the loss function.

% % === OR ====
% A large effective receptive field is essential for the understanding the global structure, and hence solving inpainting. 
% We noticed that commonly used convolutional architectures lack a sufficiently large receptive field.
% To solve this problem, we:
% Propose to build an inpainting network based on recently developed \textit{fast Fourier convolutions}~\cite{chi2020ffc}. The fast Fourier convolutions allow for the global receptive field that covers an entire image, even from the early layers of the network.
% We show that the network benefits from the fast growth of the receptive field, achieving both higher quality and parameter efficiency. Interestingly, there is some evidence that the inductive biases inside fast Fourier convolutions allow the network to generalize to high resolutions that are never seen during training (\fig{teaser}). This finding brings significant practical benefits, reducing required data and computations.

% The lack of receptive field touches not only the inpainting network, but also perceptual loss.
% We find that the use of the perceptual loss~\cite{johnson2016perceptual} based on a semantic segmentation network with a high receptive field promotes the consistency of global structures and shapes. 
\begin{figure*}[t!]
    \centering
    \includegraphics[width=1\linewidth]{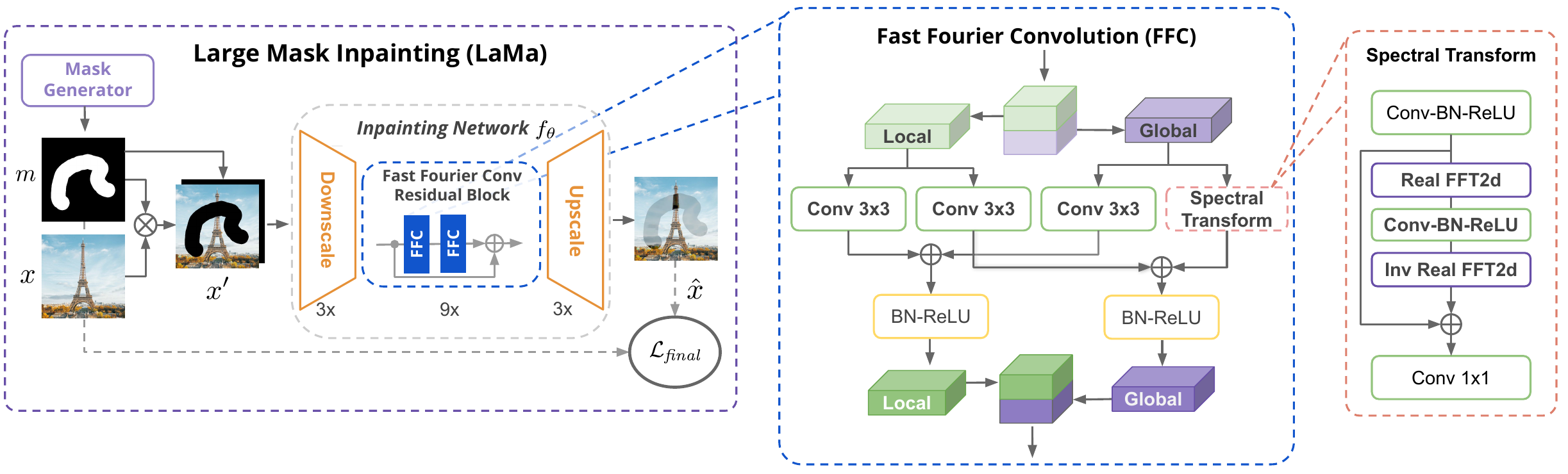}

    \caption{
    The scheme of the proposed method for large-mask inpainting (LaMa).
    LaMa is based on a feed-forward ResNet-like inpainting network that use: recently proposed fast Fourier convolution~(FFC)~\cite{chi2020ffc},
    %that have a large effective receptive field and can handle periodic structures,
    a multi-component loss that combines adversarial loss and a high receptive field perceptual loss, and a training-time large masks generation procedure.}
    \label{fig:method-scheme}
\end{figure*}

% To unlock the potential of a high receptive field, we introduce an aggressive strategy for mask generation. The procedure produces wider and bigger masks, which force the network to fully utilize the high receptive field of the model and the loss function. 

This leads us to \emph{large mask inpainting}~(LaMa)---a novel single-stage image inpainting system. The main components of LaMa are the high receptive field architecture~$(i)$, with the high receptive field loss function~$(ii)$, and the aggressive algorithm of training masks generation~$(iii)$.
We meticulously compare LaMa with state-of-the-art baselines and analyze the influence of each proposed component. Through evaluation, we find that LaMa can generalize to high-resolution images after training only on low-resolution data. LaMa can capture and generate complex periodic structures, and is robust to large masks. Furthermore, this is achieved with significantly less trainable parameters and inference time costs compared to competitive baselines. 

\section{Method}

Our goal is to inpaint a color image $x$ masked by a binary mask of unknown pixels $m$, the masked image is denoted as $x \odot m$. 
The mask $m$ is stacked with the masked image $x\odot m$, resulting in a four-channel input tensor $x'=\texttt{stack}(x\odot m,m)$. We use a feed-forward \textit{inpainting network} $f_\theta(\cdot)$, that we also refer to as generator. Taking $x'$, the inpainting network processes the input in a fully-convolutional manner, and produces an inpainted three-channel color image $\hat{x}=f_\theta(x')$ .  
The training is performed on a dataset of~(image, mask)~pairs obtained from real images and synthetically generated masks.
%\textcolor{blue}{While it is a common practice to improve the results with a 2-stage pipeline with auxiliary first-stage predictions,~e.g.,~depth~\cite{ranftl2019towards}, edges~\cite{canny1986computational}, segmentation~\cite{zhou2018semantic},  we provide a receipt that allows to a make 1-stage pipeline that in-fact can push state of the art.}

\subsection{Global context within early layers}
In challenging cases, e.g. filling of large masks, the generation of proper inpainting requires to consider global context. Thus, we argue that a good architecture should have units with as wide-as-possible receptive field as early as possible in the pipeline. The conventional fully convolutional models,~e.g. ResNet~\cite{he2016deep}, suffer from slow growth of \emph{effective receptive field}~\cite{Luo16effective}.
Receptive field might be insufficient, especially in the early layers of the network, due to the typically small (e.g. $3\times3$) convolutional kernels. 
Thus, many layers in the network will be lacking global context and will waste computations and parameters to create one.
For wide masks, the whole receptive field of a generator at the specific position may be inside the mask, thus observing only missing pixels.
The issue becomes especially pronounced for high-resolution images. 

%\vspace{5pt}
\textbf{Fast Fourier convolution~(FFC)}~\cite{chi2020ffc} is the recently proposed operator that allows to use global context in early layers. 
FFC  is based on a channel-wise fast Fourier transform~(FFT)~\cite{brigham1967fast} and has a receptive field that covers the entire image. 
FFC splits channels into two parallel branches: \emph{i)~local branch} uses conventional convolutions, and \emph{ii)~global branch} uses \textit{real FFT} to account for global context. \textit{Real FFT} can be applied only to real valued signals, and \textit{inverse real FFT} ensures that the output is real valued. \textit{Real FFT} uses only half of the spectrum compared to the FFT. 
Specifically, FFC makes following steps:
\begin{itemize}
    \setlength\itemsep{0pt}
    \setlength{\belowdisplayskip}{0pt} \setlength{\belowdisplayshortskip}{0pt}
    \setlength{\abovedisplayskip}{7pt} \setlength{\abovedisplayshortskip}{7pt}
    \item[$a)$] applies \textit{Real FFT2d} to an input tensor
    $$\textit{Real FFT2d}: \mathbb{R}^{H\times W\times C}\to\mathbb{C}^{H\times \frac{W}{2} \times C},$$
    
    \vspace{0pt}
    and concatenates real and imaginary parts
    $$\textit{ComplexToReal}: \mathbb{C}^{H\times \frac{W}{2}\times C}\to\mathbb{R}^{H\times \frac{W}{2} \times 2C};$$
    \item[$b)$] applies a convolution block in the frequency domain 
    $$\textit{ReLU}\circ\textit{BN}\circ\textit{Conv1}\!\!\times\!\!\textit{1}: \mathbb{R}^{H\times \frac{W}{2} \times 2C}\to\mathbb{R}^{H\times \frac{W}{2} \times 2C};$$
    \item[$c)$] applies inverse transform to recover a spatial structure
    $$\textit{RealToComplex}: \mathbb{R}^{H\times \frac{W}{2} \times 2C}\to\mathbb{C}^{H\times \frac{W}{2} \times C},$$
    $$\textit{Inverse Real FFT2d}: \mathbb{C}^{H\times \frac{W}{2} \times C} \to \mathbb{R}^{H\times W\times C}.$$
\end{itemize}
Finally, the outputs of the local (\emph{i}) and global (\emph{ii}) branches are fused together. The illustration of FFC is available in \fig{method-scheme}.

\textbf{The power of FFCs}~
FFCs are fully differentiable and easy-to-use drop-in replacement for conventional convolutions.
Due to the image-wide receptive field, FFCs allow the generator to account for the global context starting from the early layers, which is crucial for high-resolution image inpainting. This also leads to better efficiency: trainable parameters can be used for reasoning and generation instead of ``waiting" for a propagation of information.

We show that FFCs are well suited to capture periodic structures, which are common in human-made environments, e.g.\ bricks, ladders, windows, etc (\fig{syde-by-side-512}). 
Interestingly, sharing the same convolutions across all frequencies shifts the model towards scale equivariance~\cite{chi2020ffc} (Figures~\ref{fig:res-grow},~\ref{fig:high-res-degradation-quantitative}).

\subsection{Loss functions}

The inpainting problem is inherently ambiguous. 
There could be many plausible fillings for the same missing areas, especially when the ``holes" become wider.
We will discuss the components of the proposed loss, that together allow to handle the complex nature of the problem.

\subsubsection{High receptive field perceptual loss}
\label{hrfpl}
Naive supervised losses require the generator to reconstruct the ground truth precisely. However, the visible parts of the image often do not contain enough information for the exact reconstruction of the masked part. Therefore, using naive supervision leads to blurry results due to the averaging of multiple plausible modes of the inpainted content. 

In contrast, perceptual loss~\cite{johnson2016perceptual} evaluates a distance between features extracted from the predicted and the target images by a base pre-trained network $\phi(\cdot)$.
It does not require an exact reconstruction, allowing for variations in the reconstructed image.
The focus of large-mask inpainting is shifted towards understanding of global structure. Therefore, we argue that it is important to use the base network with a fast growth of a receptive field.
We introduce the \emph{high receptive field perceptual loss} (\textit{HRF PL}), that uses a high receptive field base model $\phi_{\text{\it HRF}}(\cdot)$:

\vspace{-17pt}
\begin{equation}
    \mathcal{L}_{\text{\it HRFPL}}(x, \hat{x}) = \mathcal{M}([\phi_{\text{\it HRF}}(x)-\phi_{\text{\it HRF}}(\hat{x})]^2),
\end{equation}

\vspace{-5pt}
\noindent where $[\cdot-\cdot]^2$ is an element-wise operation, and $\mathcal{M}$ is the sequential two-stage mean operation~(interlayer mean of intra-layer means).
The $\phi_{\text{\it HRF}}(x)$ can be implemented using Fourier or Dilated convolutions.
The HRF perceptual loss appears to be crucial for our large-mask inpainting system, as demonstrated in the ablation study (Table~\ref{tab:ablation-loss}).

\textbf{Pretext problem}~ A pretext problem on which the base network for a perceptual loss was trained is important.
For example, using a segmentation model as a backbone for perceptual loss may help to focus on high-level information, e.g. objects and their parts. On the contrary, classification models are known to focus more on textures~\cite{geirhos2018imagenettrained}, which can introduce biases harmful for high-level information.
% \textbf{Pretext problem}~ A pretext problem that was used for training of a base network for a perceptual loss is important.
% For example, using a segmentation model as a backbone for perceptual loss helps to focus on high-level information, e.g. objects and their parts. On contrary, classification models are known to focus more on textures, which is not helpful for inpainting~\cite{geirhos2018imagenettrained}. 

\subsubsection{Adversarial loss} 
We use adversarial loss to ensure that inpainting model~$f_\theta(x')$ generates naturally looking local details.
We define a discriminator $D_\xi(\cdot)$ that works on a local patch-level~\cite{isola2017image}, discriminating between ``real'' and ``fake'' patches. 
Only patches that intersect with the masked area get the ``fake'' label. 
Due to the supervised {\it HRF} perceptual loss, the generator quickly learns to copy the known parts of the input image, thus we label the known parts of generated images as ``real''.
Finally, we use the non-saturating adversarial loss:

\vspace{-12pt}
\begin{gather}
    \begin{split}
    \mathcal{L}_{\text{\textit{D}}}\!=\!-\!\mathbb{E}_{x}\Big[\log{\!D_\xi(x)}\Big]
    \!-\!\mathbb{E}_{x,m}\Big[\log{D_\xi(\hat{x})}\odot m\Big] \\ 
    \!-\!\mathbb{E}_{x,m}\Big[\log{(1\!-\!D_\xi(\hat{x}))}\odot (1-m)\Big] \\
    \end{split}
    \\
    \mathcal{L}_{\text{\textit{G}}}  = - \mathbb{E}_{x,m}\Big[\log{D_\xi(\hat{x})}\Big]\\
    \label{adv}L_{\textit{Adv}} = \texttt{sg}_\theta(\mathcal{L}_{\text{\textit{D}}}) + \texttt{sg}_\xi(\mathcal{L}_{\text{\textit{G}}}) \to \min_{\theta, \xi}
\end{gather}

\vspace{-6pt}
\noindent where $x$ is a sample from a dataset, $m$ is a synthetically generated mask, $\hat{x} = f_\theta(x')$ is the inpainting result for $x'=\texttt{stack}(x\odot m,m)$, $\texttt{sg}_\textit{var}$ stops gradients w.r.t \textit{var}, and $L_{\textit{Adv}}$ is the joint loss to optimise.

% \begin{table*}[]
% \centering
% \setlength{\tabcolsep}{0.2em}
% \resizebox{\textwidth}{!}{% <------ Don't forget this %
% \begin{tabular}{llllllllllll}
% \toprule
% {} & {\multirow[c]{2}{*}{\rotatebox[origin=c]{90}{\small  \bf \# Params}}} & \multicolumn{6}{c}{\textbf{Places} ($\bf  512 \times 512$)} & \multicolumn{4}{c}{\textbf{CelebA-HQ} ($\bf  256 \times 256$)} \\
% \cmidrule(lr){3-8}
% \cmidrule(lr){9-12}
% {} & {} & \multicolumn{2}{c}{\textbf{Narrow masks}} & \multicolumn{2}{c}{\textbf{Wide masks}} & \multicolumn{2}{c}{\textbf{Segm. masks}} & \multicolumn{2}{c}{\textbf{Narrow masks}} & \multicolumn{2}{c}{\textbf{Wide masks}} \\
% \cmidrule(lr){3-4}
% \cmidrule(lr){5-6}
% \cmidrule(lr){7-8}
% \cmidrule(lr){9-10}
% \cmidrule(lr){11-12}
% {\textbf{Method}} & {} & {FID $\downarrow$} & {LPIPS $\downarrow$} & {FID $\downarrow$} & {LPIPS $\downarrow$} & {FID $\downarrow$} & {LPIPS $\downarrow$} & {FID $\downarrow$} & {LPIPS $\downarrow$} & {FID $\downarrow$} & {LPIPS $\downarrow$} \\

\begin{table*}[]
\centering
\setlength{\tabcolsep}{0.2em}
\resizebox{\textwidth}{!}{% <------ Don't forget this %
\begin{tabular}{lrllllll@{\hskip 15pt}llll}
\toprule
{} & {\multirow[c]{3}{*}{\rotatebox[origin=c]{90}{\parbox{1.5cm}{\centering \small \bf \# Params \\ \scriptsize $\times 10^6$}}}} & \multicolumn{6}{c}{\textbf{Places} ($\bf 512 \times 512$)} & \multicolumn{4}{c}{\textbf{CelebA-HQ} ($\bf 256 \times 256$)} \\
\cmidrule(lr){3-8}
\cmidrule(lr){9-12}
{} & {} & \multicolumn{2}{c}{\textbf{Narrow masks}} & \multicolumn{2}{c}{\textbf{Wide masks}} & \multicolumn{2}{c}{\textbf{Segm. masks}} & \multicolumn{2}{c}{\textbf{Narrow masks}} & \multicolumn{2}{c}{\textbf{Wide masks}} \\
\cmidrule(lr){3-4}
\cmidrule(lr){5-6}
\cmidrule(lr){7-8}
\cmidrule(lr){9-10}
\cmidrule(lr){11-12}
{\textbf{Method}} & {} & {FID $\downarrow$} & {LPIPS $\downarrow$} & {FID $\downarrow$} & {LPIPS $\downarrow$} & {FID $\downarrow$} & {LPIPS $\downarrow$} & {FID $\downarrow$} & {LPIPS $\downarrow$} & {FID $\downarrow$} & {LPIPS $\downarrow$} \\
\midrule
LaMa-Fourier~\textcolor{trolleygrey}{(ours)} & 27$\textcolor{white}{\scriptstyle \blacktriangle}$ & $0.63$ & $0.090$ & $2.21$ & $0.135$ & $5.35$ & $0.058$ & $7.26$ & $0.085$ & $6.96$ & $0.098$ \\
\cmidrule(l){1-12}
CoModGAN~\cite{zhao2021comodgan} & 109$\textcolor{darkpink}{\scriptstyle \blacktriangle}$ & $0.82 \textcolor{darkpink}{\scriptstyle \blacktriangle30\%}$ & $0.111 \textcolor{darkpink}{\scriptstyle \blacktriangle23\%}$ & $1.82 \textcolor{darkpastelgreen}{\scriptstyle \blacktriangledown18\%}$ & $0.147 \textcolor{darkpink}{\scriptstyle \blacktriangle9\%}$ & $6.40 \textcolor{darkpink}{\scriptstyle \blacktriangle20\%}$ & $0.066 \textcolor{darkpink}{\scriptstyle \blacktriangle14\%}$ & $16.8 \textcolor{darkpink}{\scriptstyle \blacktriangle131\%}$ & $0.079 \textcolor{darkpastelgreen}{\scriptstyle \blacktriangledown7\%}$ & $24.4 \textcolor{darkpink}{\scriptstyle \blacktriangle250\%}$ & $0.102 \textcolor{darkpink}{\scriptstyle \blacktriangle4\%}$ \\
MADF~\cite{zhu2021image} & 85$\textcolor{darkpink}{\scriptstyle \blacktriangle}$ & $0.57 \textcolor{darkpastelgreen}{\scriptstyle \blacktriangledown10\%}$ & $0.085 \textcolor{darkpastelgreen}{\scriptstyle \blacktriangledown5\%}$ & $3.76 \textcolor{darkpink}{\scriptstyle \blacktriangle70\%}$ & $0.139 \textcolor{darkpink}{\scriptstyle \blacktriangle3\%}$ & $6.51 \textcolor{darkpink}{\scriptstyle \blacktriangle22\%}$ & $0.061 \textcolor{darkpink}{\scriptstyle \blacktriangle5\%}$ & --- & --- & --- & --- \\
AOT GAN~\cite{yan2021agg} & 15$\textcolor{darkpastelgreen}{\scriptstyle \blacktriangledown}$ & $0.79 \textcolor{darkpink}{\scriptstyle \blacktriangle25\%}$ & $0.091 \textcolor{darkpink}{\scriptstyle \blacktriangle1\%}$ & $5.94 \textcolor{darkpink}{\scriptstyle \blacktriangle169\%}$ & $0.149 \textcolor{darkpink}{\scriptstyle \blacktriangle11\%}$ & $7.34 \textcolor{darkpink}{\scriptstyle \blacktriangle37\%}$ & $0.063 \textcolor{darkpink}{\scriptstyle \blacktriangle10\%}$ & $6.67 \textcolor{darkpastelgreen}{\scriptstyle \blacktriangledown8\%}$ & $0.081 \textcolor{darkpastelgreen}{\scriptstyle \blacktriangledown4\%}$ & $10.3 \textcolor{darkpink}{\scriptstyle \blacktriangle48\%}$ & $0.118 \textcolor{darkpink}{\scriptstyle \blacktriangle20\%}$ \\
GCPR~\cite{hukkelaas2021image} & 30$\textcolor{darkpink}{\scriptstyle \blacktriangle}$ & $2.93 \textcolor{darkpink}{\scriptstyle \blacktriangle363\%}$ & $0.143 \textcolor{darkpink}{\scriptstyle \blacktriangle59\%}$ & $6.54 \textcolor{darkpink}{\scriptstyle \blacktriangle196\%}$ & $0.161 \textcolor{darkpink}{\scriptstyle \blacktriangle19\%}$ & $9.20 \textcolor{darkpink}{\scriptstyle \blacktriangle72\%}$ & $0.073 \textcolor{darkpink}{\scriptstyle \blacktriangle27\%}$ & --- & --- & --- & --- \\
HiFill~\cite{yi2020contextual} & 3$\textcolor{darkpastelgreen}{\scriptstyle \blacktriangledown}$ & $9.24 \textcolor{darkpink}{\scriptstyle \blacktriangle1361\%}$ & $0.218 \textcolor{darkpink}{\scriptstyle \blacktriangle142\%}$ & $12.8 \textcolor{darkpink}{\scriptstyle \blacktriangle479\%}$ & $0.180 \textcolor{darkpink}{\scriptstyle \blacktriangle34\%}$ & $12.7 \textcolor{darkpink}{\scriptstyle \blacktriangle137\%}$ & $0.085 \textcolor{darkpink}{\scriptstyle \blacktriangle49\%}$ & --- & --- & --- & --- \\
RegionWise~\cite{ma2019region} & 47$\textcolor{darkpink}{\scriptstyle \blacktriangle}$ & $0.90 \textcolor{darkpink}{\scriptstyle \blacktriangle42\%}$ & $0.102 \textcolor{darkpink}{\scriptstyle \blacktriangle14\%}$ & $4.75 \textcolor{darkpink}{\scriptstyle \blacktriangle115\%}$ & $0.149 \textcolor{darkpink}{\scriptstyle \blacktriangle11\%}$ & $7.58 \textcolor{darkpink}{\scriptstyle \blacktriangle42\%}$ & $0.066 \textcolor{darkpink}{\scriptstyle \blacktriangle14\%}$ & $11.1 \textcolor{darkpink}{\scriptstyle \blacktriangle53\%}$ & $0.124 \textcolor{darkpink}{\scriptstyle \blacktriangle46\%}$ & $8.54 \textcolor{darkpink}{\scriptstyle \blacktriangle23\%}$ & $0.121 \textcolor{darkpink}{\scriptstyle \blacktriangle23\%}$ \\
DeepFill v2~\cite{yu2019free} & 4$\textcolor{darkpastelgreen}{\scriptstyle \blacktriangledown}$ & $1.06 \textcolor{darkpink}{\scriptstyle \blacktriangle68\%}$ & $0.104 \textcolor{darkpink}{\scriptstyle \blacktriangle16\%}$ & $5.20 \textcolor{darkpink}{\scriptstyle \blacktriangle135\%}$ & $0.155 \textcolor{darkpink}{\scriptstyle \blacktriangle15\%}$ & $9.17 \textcolor{darkpink}{\scriptstyle \blacktriangle71\%}$ & $0.068 \textcolor{darkpink}{\scriptstyle \blacktriangle18\%}$ & $12.5 \textcolor{darkpink}{\scriptstyle \blacktriangle73\%}$ & $0.130 \textcolor{darkpink}{\scriptstyle \blacktriangle53\%}$ & $11.2 \textcolor{darkpink}{\scriptstyle \blacktriangle61\%}$ & $0.126 \textcolor{darkpink}{\scriptstyle \blacktriangle28\%}$ \\
EdgeConnect~\cite{nazeri2019edgeconnect} & 22$\textcolor{darkpastelgreen}{\scriptstyle \blacktriangledown}$ & $1.33 \textcolor{darkpink}{\scriptstyle \blacktriangle110\%}$ & $0.111 \textcolor{darkpink}{\scriptstyle \blacktriangle23\%}$ & $8.37 \textcolor{darkpink}{\scriptstyle \blacktriangle279\%}$ & $0.160 \textcolor{darkpink}{\scriptstyle \blacktriangle19\%}$ & $9.44 \textcolor{darkpink}{\scriptstyle \blacktriangle76\%}$ & $0.073 \textcolor{darkpink}{\scriptstyle \blacktriangle27\%}$ & $9.61 \textcolor{darkpink}{\scriptstyle \blacktriangle32\%}$ & $0.099 \textcolor{darkpink}{\scriptstyle \blacktriangle17\%}$ & $9.02 \textcolor{darkpink}{\scriptstyle \blacktriangle30\%}$ & $0.120 \textcolor{darkpink}{\scriptstyle \blacktriangle22\%}$ \\
RegionNorm~\cite{yu2020region} & 12$\textcolor{darkpastelgreen}{\scriptstyle \blacktriangledown}$ & $2.13 \textcolor{darkpink}{\scriptstyle \blacktriangle236\%}$ & $0.120 \textcolor{darkpink}{\scriptstyle \blacktriangle33\%}$ & $15.7 \textcolor{darkpink}{\scriptstyle \blacktriangle613\%}$ & $0.176 \textcolor{darkpink}{\scriptstyle \blacktriangle31\%}$ & $13.7 \textcolor{darkpink}{\scriptstyle \blacktriangle156\%}$ & $0.082 \textcolor{darkpink}{\scriptstyle \blacktriangle42\%}$ & --- & --- & --- & --- \\
\bottomrule
\end{tabular}
}
\vspace{0.2em}
\caption{
Quantitative evaluation of inpainting on Places and CelebA-HQ datasets. We report \emph{Learned perceptual image patch similarity}~(LPIPS) and  \emph{Fréchet inception distance}~(FID) metrics. 
The $\textcolor{darkpink}{\blacktriangle}$ denotes deterioration, and $\textcolor{darkpastelgreen}{\blacktriangledown}$~denotes improvement of a score compared to our LaMa-Fourier model~(presented in the first row).
The metrics are reported for different policies of test masks generation, i.e. narrow, wide, and segmentation masks. 
LaMa-Fourier consistently outperforms a wide range of the baselines.
CoModGAN~\cite{zhao2021comodgan} and MADF~\cite{zhu2021image} are the only two baselines that come close.
However, both models are much heavier than LaMa-Fourier and perform worse on average, showing that our method utilizes trainable parameters more efficiently.
}
\vspace{-10pt}
\label{tab:comparison-metrics-places}
\end{table*}

\subsubsection{The final loss function} 

In the final loss we also use $R_1\!=\!E_{x} ||\nabla D_\xi(x)||^2$ gradient penalty~\cite{Mescheder18r1,Slavin18r1,Drucker92r1}, and a \emph{discriminator-based perceptual loss} or so-called feature matching loss---a perceptual loss on the features of discriminator network~$\mathcal{L}_{\text{\it DiscPL}}$~\cite{wang2018pix2pixHD}. $\mathcal{L}_{\text{\it DiscPL}}$ is known to stabilize training, and in some cases slightly improves the performance.

The final loss function for our inpainting system 
\begin{gather}
    \mathcal{L_\textit{final}} =  \kappa L_{\textit{Adv}} + \alpha \mathcal{L}_{\text{\it HRFPL}}  + \beta\mathcal{L}_{\text{\it DiscPL}} +  \gamma R_1
\end{gather}
\noindent is the weighted sum of the discussed losses, where $L_{\textit{Adv}}$ and $\mathcal{L}_{\text{\it DiscPL}}$ are responsible for generation of naturally looking local details, while $\mathcal{L}_{\text{\it HRFPL}}$ is responsible for the supervised signal and consistency of the global structure.

\definecolor{bleudefrance}{rgb}{0.19, 0.55, 0.91}
\subsection{Generation of masks during training}
\label{mask_gen}
\begin{figure}
    \centering
    \includegraphics[width=0.45\textwidth]{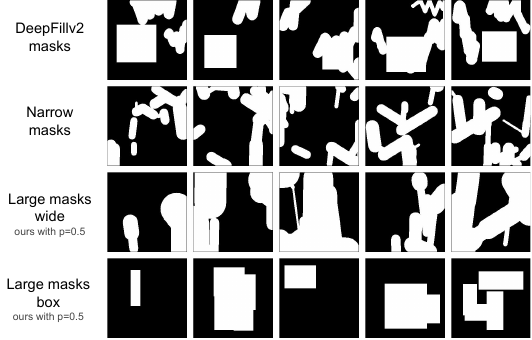}
    \caption{
    The samples from different training masks generation policies.
    We argue that the way masks are generated greatly influences the final performance of the system.
    %We argue that the way masks are generated is very important for the final performance of the system.
    Unlike the conventional practice, e.g. DeepFillv2, we use a more aggressive \textit{large mask} generation strategy where masks come uniformly either from \textit{wide masks} or \textit{box masks} strategies. 
    The masks from \textit{large mask} strategy have large area and, more importantly, are wider (see supplementary material for histograms).
    Training with our strategy helps a model to perform better on both wide and narrow masks~(Table \ref{tab:ablation-masks}). During preparation of the test datasets, we avoid masks which cover more than 50\% of an image.
    }
    \label{fig:my_label}
\end{figure}
The last component of our system is a mask generation policy.
Each training example $x'$ is a real photograph from a training dataset superimposed by a synthetically generated mask.
Similar to discriminative models where data-augmentation has a high influence on the final performance, we find that the policy of mask generation noticeably influences the performance of the inpainting system.

We thus opted for an aggressive \emph{large mask} generation strategy. 
This strategy uniformly uses samples from polygonal chains dilated by a high random width~(wide masks) and rectangles of arbitrary aspect ratios~(box masks).
The examples of our masks are demonstrated in \fig{my_label}.

We tested \emph{large mask} training against narrow mask training for several methods, and found that training with \emph{large mask} strategy generally improves performance on both narrow and wide masks (Table \ref{tab:ablation-masks}).
That suggests that increasing diversity of the masks might be beneficial for various inpainting systems.  
The sampling algorithm is provided in supplementary material.

\section{Experiments}
\vspace{-5pt}
In this section we demonstrate that the proposed technique outperforms a range of strong baselines on standard low resolutions, and the difference is even more pronounced when inpainting wider holes.
Then we conduct the ablation study, showing the importance of FFC, the high receptive field perceptual loss, and large masks.
The model, surprisingly, can generalise to high, never seen resolutions, while having significantly less parameters compared to most competitive baselines. 

\textbf{Implementation details}~For LaMa inpainting network we use a ResNet-like~\cite{he2016deep} architecture with 3 downsampling blocks, 6-18 residual blocks, and 3 upsampling blocks. 
In our model, the residual blocks use FFC.
The further details on the discriminator architecture are provided in the supplementary material.
We use Adam~\cite{kingma2014adam} optimizer, with the fixed learning rates $0.001$ and $0.0001$ for inpainting and discriminator networks, respectively.
All models are trained for 1M iterations with a batch size of 30 unless otherwise stated. 
In all experiments, we select hyperparameters using the coordinate-wise beam-search strategy.
That scheme led to the weight values $\kappa=10$, $\alpha=30$, $\beta=100$, $\gamma=0.001$.
We use these hyperparameters for the training of all models, except those described in the loss ablation study (shown in Sec.~\ref{sec:ablation-study}).
In all cases, the hyperparameter search is performed on a separate validation subset. More information about dataset splits is provided in supplementary material. 

\textbf{Data and metrics}~ We use Places~\cite{zhou2017places} and CelebA-HQ~\cite{karras2017progressive} datasets. We follow the established practice in recent image2image literature and use \emph{Learned Perceptual Image Patch Similarity}~(LPIPS)~\cite{zhang2018unreasonable} and  \emph{Fréchet inception distance}~(FID)~\cite{heusel2017gans} metrics.
Compared to pixel-level L1 and L2 distances, LPIPS and FID are more suitable for measuring performance of large masks inpainting when multiple natural completions are plausible. 
The experimentation pipeline is implemented using PyTorch~\cite{NEURIPS2019_9015}, PyTorch-Lightning~\cite{falcon2019pytorch}, and Hydra~\cite{Yadan2019Hydra}. The code and the models are publicly available at {\small \hyperref[https://github.com/saic-mdal/lama]{\tt github.com/saic-mdal/lama}}. 

\subsection{Comparisons to the baselines} 
We compare the proposed approach with a number of strong baselines that are presented in Table \ref{tab:comparison-metrics-places}. Only publicly available pretrained models are used to calculate these metrics.
For each dataset, we validate the performance across narrow, wide, and segmentation-based masks. 
% The comparison results are shown in Tables~\ref{tab:comparison-metrics-places}.  
LaMa-Fourier consistently outperforms most of the baselines, while having fewer parameters than the strongest competitors.
The only two competitive baselines CoModGAN~\cite{zhao2021comodgan} and MADF~\cite{zhu2021image} use $\approx4\times$ and $\approx 3\times$ more parameters.
The difference is especially noticeable for wide masks.

\textbf{User study~} 
% There is no perfect metric to measure the quality of generated images. 
To alleviate a possible bias of the selected metrics, we have conducted a crowdsourced user study.
The results of the user study correlate well with the quantitative evaluation and  demonstrate that the inpainting produced by our method is more preferable and less detectable compared to other methods. The protocol and the results of the user study are provided in the supplementary material. 

\begin{figure*}[t!]
\centering
    \includegraphics[width=0.98\linewidth]{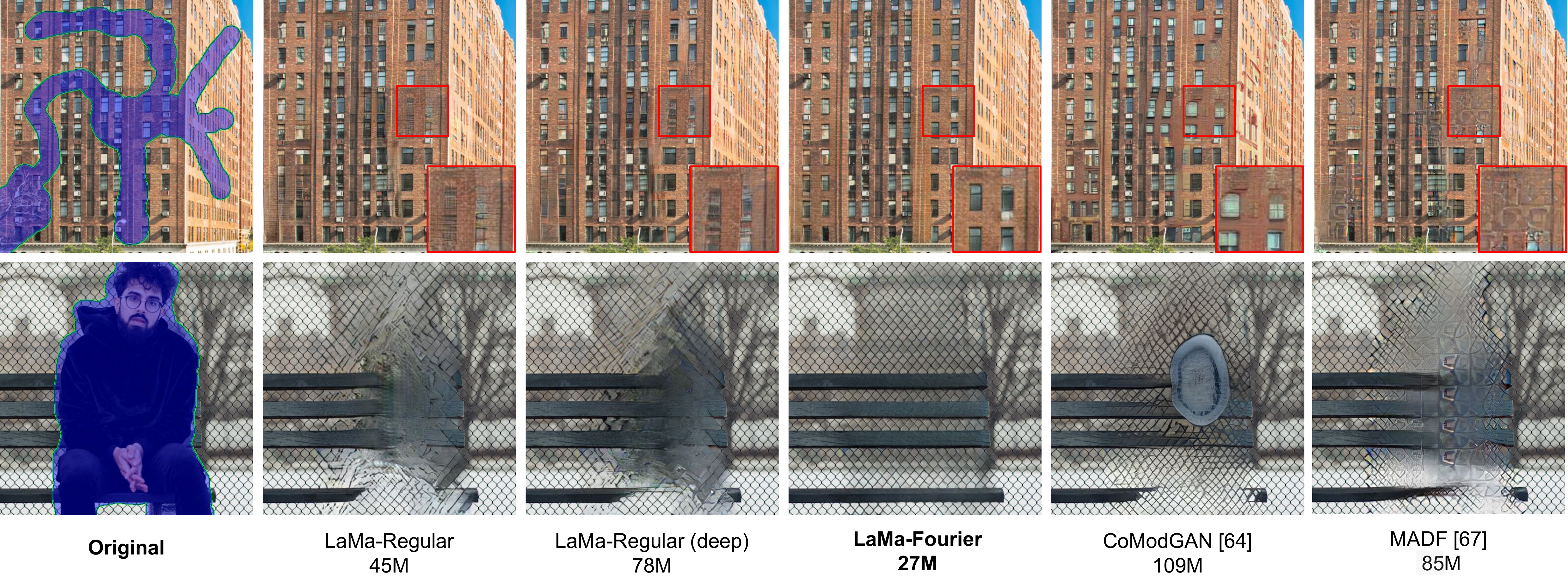}

    \caption{
The side-by-side comparison of various inpainting systems on 512$\times$512 images.
Repetitive structures, such as windows and chain-link fences are known to be hard to inpaint. FFCs allow to generate these types of structures significantly better.
%The Fourier convolutions allow to generate repetitive structures such as windows and chain-link fences---that are known to be hard to inpaint---significantly better compared to inpainting networks that are based on regular convolutions, including CoModGAN and MADF---two best-performing baselines.
Interestingly, LaMa-Fourier performs the best even with fewer parameters across the comparison while serving feasible inference time, i.e. LaMa-Fourier on average is only $\sim 20\%$ slower than LaMa-Regular.
}
    \label{fig:syde-by-side-512}
    \vspace{-0.5em}
\end{figure*}

\definecolor{bleudefrance}{rgb}{0.19, 0.55, 0.91}
\subsection{Ablation Study} 
\label{sec:ablation-study}

The goal of the study is to carefully examine the influence of different components of the method.
In this section, we present results on Places dataset; the additional results for CelebA dataset are available in supplementary material.
\begin{table}[]
\centering
\setlength{\tabcolsep}{.2em}
%%%%%%%%%%%%%%%%%%%%%%%%%%%%%%%%%%%%%%%%%%%%%%%%%%%%%%%%%%%%%%%%%%%%%%%%%%%%%
%%%%%%%%%%%%%%%%% New Autogenerated Architecture Ablation Table %%%%%%%%%%%%%
%%%%%%%%%%%%%%%%%%%%%%%%%%%%%%%%%%%%%%%%%%%%%%%%%%%%%%%%%%%%%%%%%%%%%%%%%%%%%
% \todo{show 10-30 for segm masks? now totals are shown and the difference is not significant}

\resizebox{.425\textwidth}{!}{% <------ Don't forget this %
% manual fixes
% Model to first column, second row
% \vspace{0.5em}}
% \cmidrule(l){1-8}

\begin{tabular}{llccllll}
\toprule
{} & {} & {\multirow[c]{2}{*}{\rotatebox[origin=c]{90}{\small \# Params}}} & {\multirow[c]{2}{*}{\rotatebox[origin=c]{90}{\small \# Blocks}}} & \multicolumn{2}{c}{\textbf{Narrow masks}} & \multicolumn{2}{c}{\textbf{Wide masks} \vspace{0.5em}} \\
{Model} & {Convs} & {} & {} & {FID $\downarrow$} & {LPIPS $\downarrow$} & {FID $\downarrow$ \vspace{0.3em}} & {LPIPS $\downarrow$ \vspace{0.3em}} \\
\midrule
Base & Fourier & 27 & 9 & $0.63$ & $0.090$ & $2.21$ & $0.135$ \\
\cmidrule(l){1-8}
Base & Dilated & 46 & 9 & $0.66 \textcolor{darkpink}{\scriptstyle \blacktriangle4\%}$ & $0.089 \textcolor{darkpastelgreen}{\scriptstyle \blacktriangledown1\%}$ & $2.30 \textcolor{darkpink}{\scriptstyle \blacktriangle4\%}$ & $0.136 \textcolor{darkpink}{\scriptstyle \blacktriangle1\%}$ \\
Base & Regular & 46 & 9 & $0.60 \textcolor{darkpastelgreen}{\scriptstyle \blacktriangledown5\%}$ & $0.089 \textcolor{darkpastelgreen}{\scriptstyle \blacktriangledown1\%}$ & $3.51 \textcolor{darkpink}{\scriptstyle \blacktriangle59\%}$ & $0.139 \textcolor{darkpink}{\scriptstyle \blacktriangle3\%}$ \\
Shallow & Fourier & 19 & 6 & $0.72 \textcolor{darkpink}{\scriptstyle \blacktriangle13\%}$ & $0.094 \textcolor{darkpink}{\scriptstyle \blacktriangle4\%}$ & $2.31 \textcolor{darkpink}{\scriptstyle \blacktriangle5\%}$ & $0.138 \textcolor{darkpink}{\scriptstyle \blacktriangle2\%}$ \\
Deep & Regular & 74 & 15 & $0.63 $ & $0.090 $ & $2.62 \textcolor{darkpink}{\scriptstyle \blacktriangle18\%}$ & $0.137 \textcolor{darkpink}{\scriptstyle \blacktriangle2\%}$ \\
\bottomrule
\end{tabular}

}
\caption{The table demonstrates performance of different LaMa architectures while leaving the other components the same. The $\textcolor{darkpink}{\blacktriangle}$ denotes deterioration, and $\textcolor{darkpastelgreen}{\blacktriangledown}$ denotes improvement compared to the Base-Fourier model~(presented in the first row). The FFC-based models may sacrifice a little performance on narrow masks, but significantly outperform bigger models with regular convolutions on wide masks. Visually, the FFC-based models recover complex visual structures significantly better, as shown in \fig{syde-by-side-512}.
}
\label{tab:ablation-receptive-field}
\end{table}

\vspace{-10pt}
\textbf{Receptive field of $f_\theta(\cdot)$~} FFCs increase the effective receptive field of our system. Adding FFCs substantially improves FID scores of inpainting in wide masks (Table~\ref{tab:ablation-receptive-field}).

\begin{figure*}[t!]
    \centering
    \includegraphics[width=0.95\linewidth]{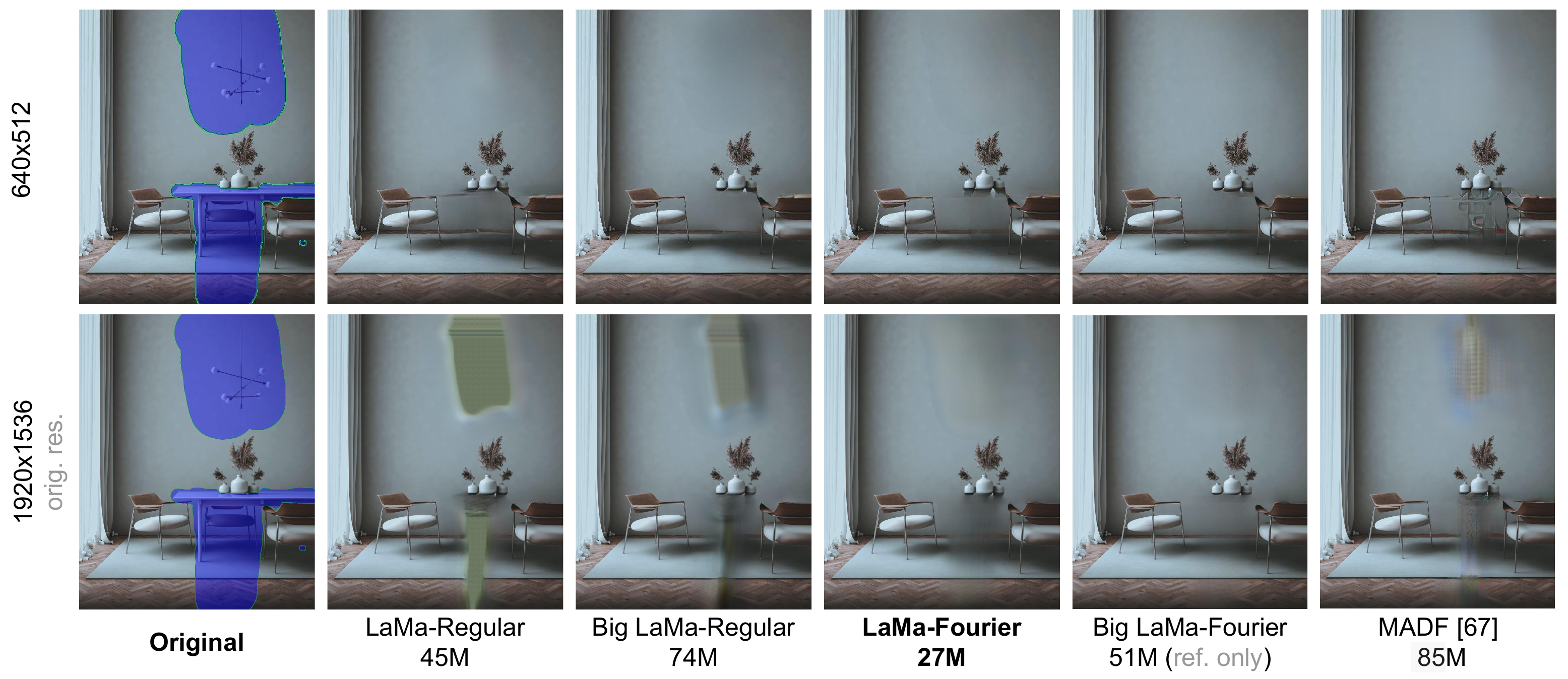}
    \vspace{-0.5em}
    \caption{
        Transfer of inpainting models to a higher resolution. All LaMa models were trained using $256\times256$ crops from $512\times512$, and MADF~\cite{zhu2021image} was trained on $512\times512$ directly.
        As the resolution increases, the models with regular convolutions swiftly start to produce critical artifacts, while FFC-based models continue to generate semantically consistent image with fine details.
        More negative and positive examples of our 51M model can be found at {\small \href{https://bit.ly/3k0gaIK}{\texttt{bit.ly/3k0gaIK}}}.
        }
    \label{fig:res-grow}
    \vspace{-1.5em}
\end{figure*}

The importance of the receptive field is most noticeable when a model is applied to a higher resolution than it was trained on.
As demonstrated in \fig{res-grow}, the model with regular convolutions produces visible artifacts as the resolution increases beyond those used at train time. The same effect is validated quantitatively (\fig{high-res-degradation-quantitative}).
FFCs also improve generation of repetitive structures such as windows a lot (\fig{syde-by-side-512}).
Interestingly, the LaMa-Fourier is only $20\%$ slower, while $40\%$ smaller than LaMa-Regular.

Dilated convolutions~\cite{yu2015multi,Chen15atrous} are an alternative option that allows the fast growth of a receptive field.
Similar to FFCs, dilated convolutions boost the performance of our inpainting system.
This further supports our hypothesis on the importance of the fast growth of the effective receptive field for image inpainting. 
However, dilated convolutions have more restrictive receptive field and heavily rely on scale, leading to inferior generalization to higher resolutions (\fig{high-res-degradation-quantitative}).
Dilated convolutions are widely implemented in most frameworks and may serve as a practical replacement for Fourier ones when the resources are limited, e.g. on mobile devices. We provide more details on the LaMa-Dilated architecture in the supplementary material.

\textbf{Loss~} 
We verify that the high receptive field of the perceptual loss---implemented with Dilated convolutions---indeed improves the quality of inpainting (\tab{ablation-loss}).
The pretext problem and the design choice beyond using dilation layers also prove to be important.
For each loss variant, we performed a weight coefficient search to ensure a fair evaluation.
%In our case, the segmentation network appeared to be a better base model for perceptual loss compared to the classification one. 
\begin{table}[]
\centering
\setlength{\tabcolsep}{0.15em}
\resizebox{0.36\textwidth}{!}{% <------ Don't forget this %
%%%%%%%%%%%%%%%%%%%%%%%%%%%%%%%%%%%%%%%%%%%%%%%%%%%%%%%%%%%%%%%%%%%%%%%%%%%%%
%%%%%%%%%%%%%%%%%%%%% New Autogenerated Loss Ablation Table %%%%%%%%%%%%%%%%%
%%%%%%%%%%%%%%%%%%%%%%%%%%%%%%%%%%%%%%%%%%%%%%%%%%%%%%%%%%%%%%%%%%%%%%%%%%%%%
% manual fixes
% Loss to first column second row
% \cmidrule(r){1-1} \cmidrule(r){2-8}
% \cmidrule(r){1-1} \cmidrule(r){2-6}
% \cmidrule(r){5-6}
\begin{tabular}{lcccll}
\toprule
{} & {} & {\multirow[c]{2}{*}{\parbox{1.05cm}{\centering \small \setstretch{0.7} Pretext \\ Problem}}} & {} & \multicolumn{2}{c}{\textbf{Segmentation masks}} \\
\cmidrule(r){5-6}
{} & {Model} & {} & {Dilation} & {FID $\downarrow$} & {LPIPS $\downarrow$} \\
\midrule
$\mathcal{L}_{\text{\it HRFPL}}$ & RN50 & Segm. & + & $5.69$ & $0.059$ \\
\cmidrule(r){1-1} \cmidrule(r){2-6}
 & RN50 & Clf. & + & $5.87 \textcolor{darkpink}{\scriptstyle \blacktriangle3\%}$ & $0.059$ \\
$\mathcal{L}_{\text{Clf\it PL}}$ & RN50 & Clf. & - & $6.00 \textcolor{darkpink}{\scriptstyle \blacktriangle5\%}$ & $0.061 \textcolor{darkpink}{\scriptstyle \blacktriangle3\%}$ \\
 & VGG19 & Clf. & - & $6.29 \textcolor{darkpink}{\scriptstyle \blacktriangle11\%}$ & $0.063 \textcolor{darkpink}{\scriptstyle \blacktriangle6\%}$ \\
\cmidrule(r){1-1} \cmidrule(r){2-6}
$\cancel{\mathcal{L}_{\text{\it PL}}}$ & - & - & - & $6.46 \textcolor{darkpink}{\scriptstyle \blacktriangle13\%}$ & $0.065 \textcolor{darkpink}{\scriptstyle \blacktriangle9\%}$ \\
\bottomrule
\end{tabular}
}
\caption{
Comparison of LaMa-Regular trained with different perceptual losses.
The $\textcolor{darkpink}{\blacktriangle}$ denotes deterioration, and~$\textcolor{darkpastelgreen}{\blacktriangledown}$ denotes improvement of a score compared to the model trained with \textit{HRF} perceptual loss based on segmentation ResNet50 with dilated convolutions~(presented in the first row).
Both dilated convolutions and pretext problem improved the scores.
}
\label{tab:ablation-loss}
\end{table}

\textbf{Masks generation~} 
Wider training masks improve inpainting of both wide and narrow holes for LaMa (ours) and RegionWise~\cite{ma2019region}  (Table~\ref{tab:ablation-masks}).
However, wider masks may make results worse, which is the case for DeepFill v2~\cite{yu2019free} and EdgeConnect~\cite{nazeri2019edgeconnect} on narrow masks.
We hypothesize that this difference is caused by specific design choices~(e.g.~high receptive field of a generator or loss functions) that make a method more or less suitable for inpainting of both narrow and wide masks at the same time.
\begin{table}[]
\centering
\setlength{\tabcolsep}{0.2em}
%%%%%%%%%%%%%%%%%%%%%%%%%%%%%%%%%%%%%%%%%%%%%%%%%%%%%%%%%%%%%%%%%%%%%%%%%%%%%
%%%%%%%%%%%%%%%%%%%% New Autogenerated Mask Ablation Table %%%%%%%%%%%%%%%%%%
%%%%%%%%%%%%%%%%%%%%%%%%%%%%%%%%%%%%%%%%%%%%%%%%%%%%%%%%%%%%%%%%%%%%%%%%%%%%%
\resizebox{0.43\textwidth}{!}{% <------ Don't forget this %
% manual fixes
% Method to third column second row
% \cmidrule(r){3-9}
\begin{tabular}{lllllll}
\toprule
{} & {} & {} & \multicolumn{2}{c}{\textbf{Narrow masks}} & \multicolumn{2}{c}{\textbf{Wide masks}} \\
{} & {} & {Method} & {FID $\downarrow$} & {LPIPS $\downarrow$} & {FID $\downarrow$} & {LPIPS $\downarrow$} \\
\midrule
\multirow[c]{8}{*}{\rotatebox[origin=c]{90}{\textsl{\textbf{Training masks}}}} & \multirow[c]{4}{*}{\rotatebox[origin=c]{90}{Narrow}} & LaMa-Regular & $0.68$ & $0.091$ & $5.41$ & $0.144$ \\
 &  & DeepFill v2 & $1.06$ & $0.104$ & $5.20$ & $0.155$ \\
 &  & EdgeConnect & $1.33$ & $0.111$ & $8.37$ & $0.160$ \\
 &  & RegionWise & $0.90$ & $0.102$ & $4.75$ & $0.149$ \\
\cmidrule(r){3-7}
 & \multirow[c]{4}{*}{\rotatebox[origin=c]{90}{Wide}} & LaMa-Regular & $0.60 \textcolor{darkpastelgreen}{\scriptstyle \blacktriangledown12\%}$ & $0.089 \textcolor{darkpastelgreen}{\scriptstyle \blacktriangledown2\%}$ & $3.51 \textcolor{darkpastelgreen}{\scriptstyle \blacktriangledown54\%}$ & $0.139 \textcolor{darkpastelgreen}{\scriptstyle \blacktriangledown4\%}$ \\
 &  & DeepFill v2 & $1.35 \textcolor{darkpink}{\scriptstyle \blacktriangle21\%}$ & $0.107 \textcolor{darkpink}{\scriptstyle \blacktriangle3\%}$ & $4.34 \textcolor{darkpastelgreen}{\scriptstyle \blacktriangledown20\%}$ & $0.148 \textcolor{darkpastelgreen}{\scriptstyle \blacktriangledown4\%}$ \\
 &  & EdgeConnect & $2.78 \textcolor{darkpink}{\scriptstyle \blacktriangle52\%}$ & $0.141 \textcolor{darkpink}{\scriptstyle \blacktriangle27\%}$ & $7.94 \textcolor{darkpastelgreen}{\scriptstyle \blacktriangledown5\%}$ & $0.160 $ \\
 &  & RegionWise & $0.74 \textcolor{darkpastelgreen}{\scriptstyle \blacktriangledown21\%}$ & $0.095 \textcolor{darkpastelgreen}{\scriptstyle \blacktriangledown7\%}$ & $3.56 \textcolor{darkpastelgreen}{\scriptstyle \blacktriangledown33\%}$ & $0.144 \textcolor{darkpastelgreen}{\scriptstyle \blacktriangledown3\%}$ \\
\bottomrule
\end{tabular}
}
\caption{
The table shows performance metrics for the training of different inpainting methods with either narrow or wide masks.
The $\textcolor{darkpink}{\blacktriangle}$ denotes deterioration, and~$\textcolor{darkpastelgreen}{\blacktriangledown}$ denotes improvement of a score induced by wide-mask training for the corresponding method.
LaMa and RegionWise inpainting clearly benefit from training with wide masks. This is an empirical evidence that the aggressive mask generation may be beneficial for other inpainting systems.
}
\label{tab:ablation-masks}
\end{table}
\subsection{Generalization to higher resolution}
Training directly at high-resolution is slow and computationally expensive.
Still, most real-world image editing scenarios require inpainting to work in high-resolution. 
So, we evaluate our models, which were trained using $256\times256$ crops from $512\times512$ images, on much larger images. 
We apply models in a fully-convolutional fashion, i.e.~an image is processed in a single pass, not patch-wise.

\begin{figure}
    \centering
    
    \includegraphics[width=0.84\linewidth]{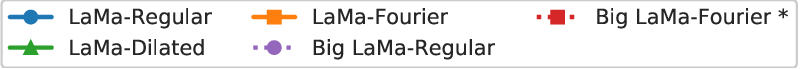}
    
    \includegraphics[width=0.84\linewidth]{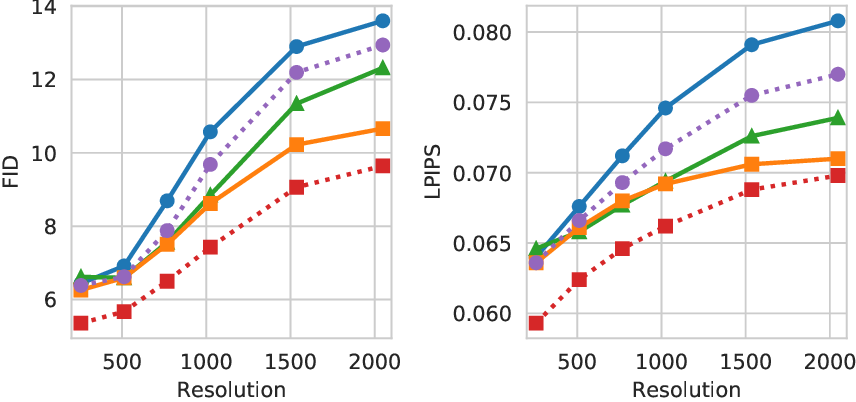}

    \caption{
    The FFC-based inpainting models can transfer to higher resolutions---that are never seen in training---with significantly smaller quality degradation. 
    All LaMa models are trained in $256\times256$ resolution.
    $^*$The Big LaMa-Fourier---our best model---is provided for reference as it was trained in different conditions (Sec. \ref{BigLaMa}).}
    \label{fig:high-res-degradation-quantitative}
\end{figure}

%\vspace{-5pt}
FFC-based models transfer to higher resolutions significantly better (\fig{high-res-degradation-quantitative}). We hypothesize that FFCs are more robust across different scales due to $i)$ image-wide receptive field, $ii)$ preserving the low-frequencies of the spectrum after scale change, $iii)$ the inherent scale equivariance of $1\times{}1$ convolutions in the frequency domain.
While all models generalize reasonably well to the $512\!\times\!512$ resolution, the FFC-enabled models preserve much more quality and consistency at the $1536\!\times\!1536$ resolution, compared to all other models (\fig{res-grow}). 
It is worth noting, that they achieve this quality at a significantly lower parameter cost than the competitive baselines.

%\vspace{-5pt}
\subsection{Teaser model: Big LaMa}
\label{BigLaMa}
To verify the scalability and applicability of our approach to real high-resolution images,
we trained a large inpainting Big LaMa model with more resources.

Big LaMa-Fourier differs from LaMa-Fourier in three aspects: the depth of the generator; the training dataset; and the size of the batch. It has 18 residual blocks, all based on FFC, resulting in 51M parameters.
The model was trained on a subset of 4.5M images from Places-Challenge dataset~\cite{zhou2017places}.
Just as our standard base model, the Big LaMa was trained only on low-resolution $256\times256$ crops of approximately $512\times512$ images. Big LaMa uses a larger batch size of 120 (instead of 30 for our other models).
Although we consider this model relatively large, it is still smaller than some of the baselines. It was trained on eight NVidia V100 GPUs for approximately 240 hours.
The inpainting examples of Big LaMa model are presented in Figures \ref{fig:teaser} and \ref{fig:res-grow}.

\section{Related Work}

Early data-driven approaches to image inpainting relied on patch-based~\cite{Criminisi03exemplar} and nearest neighbor-based~\cite{Hays07millions} generation.
%Advances in deep learning have made data-driven approaches to image inpainting considerably more popular.
%Thus,~\cite{pathak2016context} was among the first works that used a convolutional neural network with an encoder-decoder architecture trained in an adversarial way~\cite{goodfellow2014generative}.
One of the first inpainting works in deep learning era~\cite{pathak2016context} used a convnet with an encoder-decoder architecture trained in an adversarial way~\cite{goodfellow2014generative}.
This approach remains commonly used for deep inpainting to date.
Another popular group of choices for the completion network is architectures based on U-Net~\cite{ronneberger2015u}, such as~\cite{liu2018image, yan2018shift, zeng2019learning, liu2020rethinking}.

One common concern is the ability of the network to grasp the local and global context. Towards this end,~\cite{iizuka2017globally} proposed to incorporate dilated convolutions~\cite{yu2015multi} to expand receptive field; besides, two discriminators were supposed to encourage global and local consistency separately. In~\cite{wang2018image}, the use of branches in the completion network with varying receptive fields was suggested. To borrow information from spatially distant patches,~\cite{yu2018generative} proposed the contextual attention layer. Alternative attention mechanisms were suggested in~\cite{liu2019coherent, xie2019image, zheng2019pluralistic}. Our study confirms the importance of the efficient propagation of information between distant locations.
%One variant of our approach relies heavily on dilated convolutional blocks, as proposed recently for matching tasks in~\cite{schuster2019sdc}.
One variant of our approach relies heavily on dilated convolutional blocks, inspired by~\cite{schuster2019sdc}.
As an even better alternative, we propose a mechanism based on transformations in the frequency domain (FFC)~\cite{chi2020ffc}. This also aligns with a recent trend on using Transformers in computer vision~\cite{dosovitskiy2020image,esser2021taming} and treating Fourier transform as a lightweight replacement to the self-attention~\cite{lee2021fnet,rao2021global}.

At a more global level,~\cite{yu2018generative} introduced a coarse-to-fine framework that involves two networks. In their approach, the first network completes coarse global structure in the holes, while the second network then uses it as a guidance to refine local details. Such two-stage approaches that follow a relatively old idea of structure-texture decomposition~\cite{Bertalmio03texturestructure} became prevalent in the subsequent works. Some studies~\cite{sagong2019pepsi, shin2020pepsi++} modify the framework so that coarse and fine result components are obtained simultaneously rather than sequentially. Several works suggest two-stage methods that use completion of other structure types as an intermediate step: salient edges in~\cite{nazeri2019edgeconnect}, semantic segmentation maps in~\cite{song2018spg}, foreground object contours in~\cite{xiong2019foreground}, gradient maps in~\cite{yang2020learning}, and edge-preserved smooth images in~\cite{ren2019structureflow}. Another trend is progressive approaches~\cite{zhang2018semantic, guo2019progressive, li2020recurrent, zeng2020high}.
%In contrast to all these recent works, our study effectively advocates the use of a simpler single-stage system. We show that with proper design choices the single-stage approach leads to very strong results.
In contrast to all these works, we demonstrate that a meticulously designed single-stage approach can achieve very strong results.

To deal with irregular masks, several works modified convolutional layers, introducing partial~\cite{liu2018image}, gated~\cite{yu2019free}, light-weight gated~\cite{yi2020contextual} and region-wise~\cite{ma2019region} convolutions.
%Other works considered the generation of masks for training the inpainting networks, including random holes~\cite{iizuka2017globally}, free-form strokes~\cite{yu2019free} and object-shaped masks~\cite{yi2020contextual, zeng2020high}.
Various shapes of training masks were explored, including random~\cite{iizuka2017globally}, free-form~\cite{yu2019free} and object-shaped masks~\cite{yi2020contextual, zeng2020high}.
We found that as long as contours of training masks are diverse enough, the exact way of mask generation is not as important as the width of the masks.

Many losses were proposed to train inpainting networks. Typically, pixel-wise (e.g. $\ell1$, $\ell2$) and adversarial losses are used. Some approaches apply spatially discounted weighting strategies for a pixel-wise loss~\cite{pathak2016context, yeh2017semantic, yu2018generative}. Simple convolutional discriminators~\cite{pathak2016context, yang2020learning} or PatchGAN discriminators~\cite{iizuka2017globally, zeng2019learning, ren2019structureflow, liu2019coherent} were used to implement adversarial losses. Other popular choices are Wasserstein adversarial losses with gradient-penalized discriminators~\cite{yu2018generative, yi2020contextual} and spectral-normalized discriminators~\cite{nazeri2019edgeconnect, yu2019free, liu2020rethinking, zeng2020high}. Following previous works~\cite{Mescheder18r1,karras2019style}, we use an r1-gradient penalized patch discriminator in our system. A perceptual loss is also commonly applied, usually with VGG-16~\cite{liu2018image, xie2019image, li2020recurrent, liu2020rethinking} or VGG-19~\cite{yang2017high, song2018contextual, nazeri2019edgeconnect, yang2020learning} backbones pretrained on ImageNet classification~\cite{russakovsky2015imagenet}. In contrast to those works, we have found that such perceptual losses are suboptimal for image inpainting and proposed a better alternative.
% pretrained network for perceptual feature extraction. 
Inpainting frameworks often incorporate style~\cite{liu2018image, ma2019region, ma2019region, xie2019image, nazeri2019edgeconnect, li2020recurrent} and feature matching~\cite{yang2017high, song2018spg, nazeri2019edgeconnect, hui2020image} losses. The latter is also employed in our system.

\section{Discussion}

\vspace{-0.5em}
In this study, we have investigated the use of a simple, single-stage approach for large-mask inpainting. We have shown that such an approach is very competitive and can push the state of the art in image inpainting, given the appropriate choices of the architecture, the loss function, and the mask generation strategy. 
The proposed method is arguably good in generating repetitive visual structures~(\fig{teaser},~\ref{fig:syde-by-side-512}), which appears to be an issue for many inpainting methods. 
However, LaMa usually struggles when a strong perspective distortion gets involved~(see supplementary material).
We would like to note that this is usually the case for complex images from the Internet, that do not belong to a dataset.
It remains a question whether FFCs can account for these deformations of periodic signals.
Interestingly, FFCs allow the method to generalize to never seen high resolutions, and be more parameter-efficient compared to state-of-the-art baselines.
The Fourier or Dilated convolutions are not the only options to receive a high receptive field. For instance, a high receptive field can be obtained with vision transformer \cite{dosovitskiy2020image} that is also an exciting topic for future research. 
We believe that models with a large receptive field will open new opportunities for the development of efficient high-resolution computer vision models.

\vspace{3pt}
\noindent\textbf{Acknowledgements~} We want to thank Nikita Dvornik, Gleb Sterkin, Aibek Alanov, Anna Vorontsova, Alexander Grishin, and Julia Churkina for their valuable feedback. 

\vspace{3pt}
\noindent\textbf{Supplementary material~} For more details and visual samples, please refer to the project page {\small \url{https://saic-mdal.github.io/lama-project/}} or supplementary material {\small \url{https://bit.ly/3zhv2rD}}.

\FloatBarrier
% \ifnum\value{page}>8 \todo{Number of pages exceeded!!!!}\fi

{\small
\bibliographystyle{ieee_fullname}
\bibliography{refs}
}
\end{document}